\documentclass[11pt,a4paper]{article}
\usepackage[hyperref]{naaclhlt2019}
\usepackage{times}
\usepackage{latexsym}
\usepackage{epsfig}
\usepackage{graphicx}

\usepackage{xcolor}
\definecolor{dark_red}{rgb}{0.9,0.0,0.0}
\definecolor{dark_green}{rgb}{0.0,0.6,0.0}
\definecolor{dark_blue}{rgb}{0.0,0.0,0.9}

\usepackage{url}

\usepackage{amsmath}
\usepackage{amssymb}

\aclfinalcopy 
\def\aclpaperid{173} 

\title{Vector of Locally-Aggregated Word Embeddings (VLAWE):\\
A Novel Document-level Representation}

\author{{Radu Tudor} Ionescu \and Andrei M. Butnaru\\
  University of Bucharest\\
  Department of Computer Science\\
  14 Academiei, Bucharest, Romania\\
  {\tt raducu.ionescu@gmail.com}\\
  {\tt butnaruandreimadalin@gmail.com}
}

\date{}

\begin{document}
\maketitle
\begin{abstract}
In this paper, we propose a novel representation for text documents based on aggregating word embedding vectors into document embeddings. Our approach is inspired by the Vector of Locally-Aggregated Descriptors used for image representation, and it works as follows. First, the word embeddings gathered from a collection of documents are clustered by k-means in order to learn a codebook of semnatically-related word embeddings. Each word embedding is then associated to its nearest cluster centroid (codeword). The Vector of Locally-Aggregated Word Embeddings (VLAWE) representation of a document is then computed by accumulating the differences between each codeword vector and each word vector (from the document) associated to the respective codeword. We plug the VLAWE representation, which is learned in an unsupervised manner, into a classifier and show that it is useful for a diverse set of text classification tasks. We compare our approach with a broad range of recent state-of-the-art methods, demonstrating the effectiveness of our approach. Furthermore, we obtain a considerable improvement on the Movie Review data set, reporting an accuracy of $93.3\%$, which represents an absolute gain of $10\%$ over the state-of-the-art approach. Our code is available at https://github.com/raduionescu/vlawe-boswe/.
\end{abstract}

\setlength{\abovedisplayskip}{4pt}
\setlength{\belowdisplayskip}{4pt}

\section{Introduction}
\vspace*{-0.1cm}

In recent years, word embeddings \cite{Bengio-JMLR-2003,Collobert-ICML-2008,Mikolov-NIPS-2013,Pennington-EMNLP-2014} have had a huge impact in natural language processing (NLP) and related fields, being used in many tasks including sentiment analysis \cite{Dos-COLING-2014,Fu-ESA-2018}, information retrieval \cite{Perronnin-ACL-2013,Ye-ACM-2016} and word sense disambiguation \cite{Bhingardive-NAACL-2015,Butnaru-EACL-2017,Chen-EMNLP-2014,Navigli-ACL-2016}, among many others. Starting from word embeddings, researchers proposed various ways of aggregating word embedding vectors to obtain efficient sentence-level or document-level representations \cite{Ionescu-KES-2017,Cheng-IJCAI-2018,Perronnin-ACL-2013,Conneau-EMNLP-2017,Cozma-ACL-2018,Fu-ESA-2018,Hill-NAACL-2016,Kiros-NIPS-2015,Kusner-ICML-2015,Le-ICML-2014,Shen-ACL-2018,Torki-ACL-2018,Zhao-IJCAI-2015,Zhou-COLING-2016,Zhou-IJCAI-2018}. Although the mean (or sum) of word vectors is commonly adopted because of its simplicity \cite{Mitchell-CS-2010}, it seems that more complex approaches usually yield better performance \cite{Cheng-IJCAI-2018,Conneau-EMNLP-2017,Cozma-ACL-2018,Fu-ESA-2018,Hill-NAACL-2016,Kiros-NIPS-2015,Torki-ACL-2018,Zhao-IJCAI-2015,Zhou-COLING-2016,Zhou-IJCAI-2018}. To this end, we propose a simple yet effective approach for aggregating word embeddings into document embeddings. Our approach is inspired by the Vector of Locally-Aggregated Descriptors (VLAD) \cite{Jegou-CVPR-2010,Jegou-PAMI-2012} used in computer vision to efficiently represent images for various image classification and retrieval tasks. To our knowledge, we are the first to adapt and use VLAD in the text domain.

Our document-level representation is constructed as follows. First, we apply a pre-trained word embedding model, such as \emph{GloVe} \cite{Pennington-EMNLP-2014}, on all the words from a set of training documents in order to obtain a set of training word vectors. The word vectors are clustered by k-means in order to learn a codebook of semnatically-related word embeddings. Each word embedding is then associated to its nearest cluster centroid (codeword). The Vector of Locally-Aggregated Word Embeddings (VLAWE) representation of a text document is then computed by accumulating the differences between each codeword vector and each word vector that is both present in the document and associated to the respective codeword. Since our approach considers cluster centroids as reference for building the representation, it can easily accommodate new words, not seen during k-means training, simply by associating them to the nearest cluster centroids. Thus, VLAWE is robust to vocabulary distribution gaps between training and test, which can appear when the training set is particularly smaller or from a different domain. Certainly, the robustness holds as long as the word embeddings are pre-trained on a very large set of documents, e.g. the entire Wikipedia.

We plug the VLAWE representation, which is learned in an unsupervised manner, into a classifier, namely Support Vector Machines (SVM), and show that it is useful for a diverse set of text classification tasks. We consider five benchmark data sets: Reuters-21578 \cite{reuters-21578}, RT-2k \cite{Pang-ACL-2004}, MR \cite{Pang-ACL-2005}, TREC \cite{Li-COLING-2002} and Subj \cite{Pang-ACL-2004}. We compare VLAWE with recent state-of-the-art methods \cite{Ionescu-KES-2017,Cheng-IJCAI-2018,Fu-ESA-2018,Hill-NAACL-2016,Iyyer-ACL-2015,Kim-EMNLP-2014,Kiros-NIPS-2015,Le-ICML-2014,Liu-IJCAI-2017,Shen-ACL-2018,Torki-ACL-2018,Xue-TKDE-2009,Zhao-IJCAI-2015,Zhou-COLING-2016,Zhou-IJCAI-2018}, demonstrating the effectiveness of our approach. Furthermore, we obtain a considerable improvement on the Movie Review (MR) data set, surpassing the state-of-the-art approach of \newcite{Cheng-IJCAI-2018} by almost $10\%$.

The rest of the paper is organized as follows. We present related works on learning document-level representations in Section~\ref{sec:RW}. We describe the Vector of Locally-Aggregated Word Embeddings in Section~\ref{sec:M}. We present experiments and results on various text classification tasks in Section~\ref{sec:E}. Finally, we draw our conclusion in Section~\ref{sec:C}.

\vspace*{-0.1cm}
\section{Related Work}
\label{sec:RW}
\vspace*{-0.1cm}

There are various works \cite{Ionescu-KES-2017,Cheng-IJCAI-2018,Conneau-EMNLP-2017,Fu-ESA-2018,Hill-NAACL-2016,Iyyer-ACL-2015,Kim-EMNLP-2014,Kiros-NIPS-2015,Kusner-ICML-2015,Le-ICML-2014,Perronnin-ACL-2013,Shen-ACL-2018,Torki-ACL-2018,Zhao-IJCAI-2015,Zhou-IJCAI-2018} that propose to build effective sentence-level or document-level representations based on word embeddings. While most of these approaches are based on deep learning \cite{Cheng-IJCAI-2018,Conneau-EMNLP-2017,Hill-NAACL-2016,Iyyer-ACL-2015,Kim-EMNLP-2014,Kiros-NIPS-2015,Le-ICML-2014,Zhao-IJCAI-2015,Zhou-IJCAI-2018}, there have been some approaches that are inspired by computer vision research, namely by the bag-of-visual-words \cite{Ionescu-KES-2017} and by Fisher Vectors \cite{Perronnin-ACL-2013}. The relationship between the bag-of-visual-words, Fisher Vectors and VLAD is discussed in \cite{Jegou-PAMI-2012}. The discussion can be transferred to describe the relantionship of our work and the closely-related works of \newcite{Ionescu-KES-2017} and \newcite{Perronnin-ACL-2013}.

\vspace*{-0.1cm}
\section{Method}
\label{sec:M}
\vspace*{-0.1cm}

The Vector of Locally-Aggregated Descriptors (VLAD) \cite{Jegou-CVPR-2010,Jegou-PAMI-2012} was introduced in computer vision to efficiently represent images for various image classification and retrieval tasks. We propose to adapt the VLAD representation in order to represent text documents instead of images. Our adaptation consists of replacing the Scale-Invariant Feature Transform (SIFT) image descriptors \cite{Lowe-SIFT-2004} useful for recognizing object patterns in images with word embeddings \cite{Mikolov-NIPS-2013,Pennington-EMNLP-2014} useful for recognizing semantic patterns in text documents. We coin the term \emph{Vector of Locally-Aggregated Word Embeddings (VLAWE)} for the resulting document representation. 

The VLAWE representation is derived as follows. First, each word in the collection of training documents is represented as a word vector using a pre-trained word embeddings model. The result is a set $X = \{x_1, x_2, ..., x_n\}$ of $n$ word vectors. As for the VLAD model, the next step is to learn a codebook $\{\mu_1, \mu_2, ...,\mu_k\}$ of representative meta-word vectors (codewords) using k-means. Each codeword $\mu_i$ is the centroid of the cluster $C_i \subset X$:
\begin{equation}
\mu_i = \frac{1}{|C_i|} \sum_{x_t \in C_i} x_t, \forall i \in \{1,2, ...,k\},
\end{equation}
where $|C_i|$ is the number of word vectors assigned to cluster $C_i$ and $k$ is the number of clusters. Since word embeddings carry semantic information by projecting semantically-related words in the same region of the embedding space, it means that the resulting clusters contain semantically-related words. The formed centroids are stored in a randomized forest of k-d trees to reduce search cost, as described in \cite{Philbin-2007,Ionescu-WREPL-2013,Ionescu-ICIP-2014,Ionescu-ICIAP-2015}. Each word embedding $x_t$ is associated to a single cluster $C_i$, such that the Euclidean distance between $x_t$ and the corresponding codeword $\mu_i$ is minimum, for all $i \in \{1, 2, ..., k\}$. For each document $D$ and each codeword $\mu_i$, the differences $x_t - \mu_i$ of the vectors $x_t \in C_i \cap D $ and the codeword $\mu_i$ are accumulated into column vectors:
\begin{equation}
v_{i,D} = \sum_{x_t \in C_i \cap D} x_t - \mu_i,
\end{equation}
where $D \subset X$ is the set of word embeddings in a given text document. The final VLAWE embedding for a given document $D$ is obtained by stacking together the $d$-dimensional residual vectors $v_{i,D}$, where $d$ is equal to the dimension of the word embeddings:
\begin{equation}
\begin{split}
\phi_D = \begin{bmatrix}
           v_{1,D} \\
           v_{2,D} \\
           \vdots \\
           v_{k,D}
         \end{bmatrix}.
\end{split}
\end{equation}
Therefore, the VLAWE document embedding is has $k \cdot d$ components. 

The VLAWE vector $\phi_D$ undergoes two normalization steps. First, a power normalization is performed by applying the following operator independently on each component (element):
\begin{equation}\label{eq_pownorm}
f(z) = sign(z) \cdot |z|^{\alpha},
\end{equation}
where $0 \leq \alpha \leq 1$ and $|z|$ is the absolute value of $z$. Since words in natural language follow the Zipf's law~\cite{Powers-ACL-1998}, it seems natural to apply the power normalization in order to reduce the influence of highly frequent words, e.g. common words or stopwords, which can corrupt the representation. As \newcite{Jegou-PAMI-2012}, we empirically observed that this step consistently improves the quality of the representation. The power normalized document embeddings are then $L_2$-normalized. 
After obtaining the normalized VLAWE representations, we employ a classification method to learn a discriminative model for each specific text classification task.

\vspace*{-0.1cm}
\section{Experiments}
\label{sec:E}
\vspace*{-0.1cm}
\subsection{Data Sets}
\vspace*{-0.1cm}

We exhibit the performance of VLAWE on five public data sets: Reuters-21578 \cite{reuters-21578}, RT-2k \cite{Pang-ACL-2004}, MR \cite{Pang-ACL-2005}, TREC \cite{Li-COLING-2002} and Subj \cite{Pang-ACL-2004}.

The Reuters-21578 data set~\cite{reuters-21578} contains articles collected from Reuters newswire. Following \newcite{Joachims-ECML-1998} and \newcite{Yang-SIGIR-1999}, we select the categories (topics) that have at least one document in the training set and one in the test set, leading to a total of $90$ categories. We use the ModeApte evaluation \cite{Xue-TKDE-2009}, in which unlabeled documents are eliminated, leaving a total of $10787$ documents. The collection is already divided into $7768$ documents for training and $3019$ documents for testing. 

The RT-2k data set \cite{Pang-ACL-2004} consists of $2000$ movie reviews taken from the IMDB movie review archives. There are $1000$ positive reviews rated with four or five stars, and $1000$ negative reviews rated with one or two stars. The task is to discriminate between positive and negative reviews.

The Movie Review (MR) data set \cite{Pang-ACL-2005} consists of $5331$ positive and $5331$ negative sentences. Each sentence is selected from one movie review. The task is to discriminate between positive and negative sentiment.

TREC \cite{Li-COLING-2002} is a question type classification data set, where questions are divided into $6$ classes. The collection is already divided into $5452$ questions for training and $500$ questions for testing.

The Subjectivity (Subj) \cite{Pang-ACL-2004} data set contains $5000$ objective and $5000$ subjective sentences. The task is to classify a sentence as being either subjective or objective.

\begin{table*}[!t]
\setlength\tabcolsep{5.5pt}
\begin{center}
\begin{tabular}{lccccc}
\hline
Method 											& Reuters-21578 	& RT-2k				&	MR					& TREC					& Subj \\
\hline
\hline
Average of word embeddings (baseline) 	& $85.3$ 		& $84.7$ 				& $77.4$				& $80.0$				& $89.5$\\
BOW (baseline) 								& $86.5$ 				& $84.1$ 				& $77.1$				& $89.3$				& $89.3$\\
\hline
TF + FA + CP + SVM \cite{Xue-TKDE-2009}	& $\textcolor{dark_blue}{87.0}$	& - 	& - 						& -						& -\\
														
Paragraph vectors \cite{Le-ICML-2014}		& - 				& - 						& $74.8$ 				& $91.8$				& $90.5$\\
															
CNN \cite{Kim-EMNLP-2014}			& - 						& $83.5$ 				& $81.5$				& $93.6$ 				& $93.4$\\

DAN \cite{Iyyer-ACL-2015}						& - 				& - 						& $80.1$ 				& -						& -\\

Combine-skip \cite{Kiros-NIPS-2015}	& - 					& - 						& $76.5$ 				& $92.2$				& $93.6$\\

Combine-skip + NB \cite{Kiros-NIPS-2015}	& - 			& - 						& $80.4$ 				& -						& $93.6$\\

AdaSent \cite{Zhao-IJCAI-2015}		& -						& - 						& $83.1$ 				& $92.4$				& $\textcolor{dark_green}{95.5}$\\

SAE + embs. \cite{Hill-NAACL-2016}		& - 					& - 						& $73.2$ 				& $80.4$				& $89.8$\\

SDAE + embs. \cite{Hill-NAACL-2016}	& - 					& - 						& $74.6$ 				& $78.4$				& $90.8$\\

FastSent + AE \cite{Hill-NAACL-2016}	& - 					& - 						& $71.8$ 				& $80.4$				& $88.8$\\

BLSTM \cite{Zhou-COLING-2016}					& - 			& -						& $80.0$ 				& $93.0$				& $92.1$\\

BLSTM-Att \cite{Zhou-COLING-2016}			& - 			& - 						& $81.0$ 				& $93.8$				& $93.5$\\

BLSTM-2DCNN \cite{Zhou-COLING-2016}	& - 			& - 						& $82.3$ 				& $\textcolor{dark_red}{96.1}$	& $94.0$\\

DC-TreeLSTM \cite{Liu-IJCAI-2017}				& - 			& - 						& $81.7$ 				& $93.8$				& $93.7$\\

BOSWE \cite{Ionescu-KES-2017}		& $\textcolor{dark_green}{87.2}$ 	& $\textcolor{dark_blue}{89.7}$ 	& - 	& -						& -\\

TreeNet \cite{Cheng-IJCAI-2018}		& - 						& -			 			& $79.8$ 				& $91.6$				& $92.0$\\

TreeNet-GloVe \cite{Cheng-IJCAI-2018}	& - 				& -		& $\textcolor{dark_green}{83.6}$ 	& $\textcolor{dark_red}{96.1}$	& $\textcolor{dark_red}{95.9}$\\

BOMV \cite{Fu-ESA-2018}					& - 						& $\textcolor{dark_green}{90.2}$ 	& - 			& -						& $90.9$\\

SWEM-average \cite{Shen-ACL-2018}    & - 					& -			 			& $77.6$				& $92.2$				& $92.5$\\

SWEM-concat \cite{Shen-ACL-2018}	& - 					& -			 			& $78.2$ 				& $91.8$				& $93.0$\\

COV + Mean \cite{Torki-ACL-2018}	& - 						& -			 			& $80.2$ 				& $90.3$				& $93.1$\\

COV + BOW \cite{Torki-ACL-2018}	& - 						& -			 			& $80.7$ 				& $91.8$				& $93.3$\\

COV + Mean + BOW \cite{Torki-ACL-2018}	& - 			& - 						& $81.1$ 				& $91.6$				& $93.2$\\

DARLM \cite{Zhou-IJCAI-2018}			& - 						& -			 	& $\textcolor{dark_blue}{83.2}$ 	& $\textcolor{dark_green}{96.0}$	& $94.1$\\
										
\hline
VLAWE (ours)	& $\textcolor{dark_red}{89.3}$ 	& $\textcolor{dark_red}{94.1}$ 	& $\textcolor{dark_red}{93.3}$ 	& $\textcolor{dark_blue}{94.2}$	& $\textcolor{dark_blue}{95.0}$\\
\hline
\end{tabular}
\end{center}
\vspace*{-0.2cm}
\caption{Performance results (in $\%$) of our approach (VLAWE) versus several state-of-the-art methods \cite{Ionescu-KES-2017,Cheng-IJCAI-2018,Fu-ESA-2018,Hill-NAACL-2016,Iyyer-ACL-2015,Kim-EMNLP-2014,Kiros-NIPS-2015,Le-ICML-2014,Liu-IJCAI-2017,Shen-ACL-2018,Torki-ACL-2018,Xue-TKDE-2009,Zhao-IJCAI-2015,Zhou-COLING-2016,Zhou-IJCAI-2018} on the Reuters-21578, RT-2k, MR, TREC and Subj data sets. The top three results on each data set are highlighted in \textcolor{dark_red}{red}, \textcolor{dark_green}{green} and \textcolor{dark_blue}{blue}, respectively. Best viewed in color.}
\label{tab_VLAWE}
\vspace*{-0.3cm}
\end{table*}

\vspace*{-0.1cm}
\subsection{Evaluation and Implementation Details}
\vspace*{-0.1cm}

In the experiments, we used the pre-trained word embeddings computed with the \emph{GloVe} toolkit provided by \newcite{Pennington-EMNLP-2014}. The pre-trained GloVe model contains $300$-dimensional vectors for $2.2$ million words and phrases. Most of the steps required for building the VLAWE representation, such as the k-means clustering and the randomized forest of k-d trees, are implemented using the VLFeat library~\cite{vedaldi-vlfeat-2008}. We set the number of clusters (size of the codebook) to $k = 10$, leading to a VLAWE representation of $k \cdot d = 10 \cdot 300 = 3000$ components. Similar to \newcite{Jegou-PAMI-2012}, we set $\alpha = 0.5$ for the power normalization step in Equation~\eqref{eq_pownorm}, which consistently leads to near-optimal results on all data sets. In the learning stage, we employ the Support Vector Machines (SVM) implementation provided by LibSVM \cite{LibSVM-2011}. We set the SVM regularization parameter to $C=1$ in all our experiments. In the SVM, we use the linear kernel. For optimal results, the VLAWE representation is combined with the BOSWE representation \cite{Ionescu-KES-2017}, which is based on the PQ kernel \cite{Ionescu-ICIAP-2013a,Ionescu-PRL-2015}. 

We follow the same evaluation procedure as \newcite{Kiros-NIPS-2015} and \newcite{Hill-NAACL-2016}, using $10$-fold cross-validation when a train and test split is not pre-defined for a given data set. As evaluation metrics, we employ the micro-averaged $F_1$ measure for the Reuters-21578 data set and the standard classification accuracy for the RT-2k, the MR, the TREC and the Subj data sets, in order to fairly compare with the related art.

\vspace*{-0.1cm}
\subsection{Results}
\vspace*{-0.1cm}

We compare VLAWE with several state-of-the-art methods \cite{Ionescu-KES-2017,Cheng-IJCAI-2018,Fu-ESA-2018,Hill-NAACL-2016,Iyyer-ACL-2015,Kim-EMNLP-2014,Kiros-NIPS-2015,Le-ICML-2014,Liu-IJCAI-2017,Shen-ACL-2018,Torki-ACL-2018,Xue-TKDE-2009,Zhao-IJCAI-2015,Zhou-COLING-2016,Zhou-IJCAI-2018} as well as two baseline methods, namely the average of word embeddings and the standard bag-of-words (BOW). The corresponding results are presented in Table \ref{tab_VLAWE}.

First, we notice that our approach outperforms both baselines on all data sets, unlike other related methods \cite{Le-ICML-2014,Hill-NAACL-2016}. In most cases, our improvements over the baselines are higher than $5\%$. On the Reuters-21578 data set, we surpass the closely-related approach of \newcite{Ionescu-KES-2017} by around $2\%$. On the RT-2k data set, we surpass the related works of \newcite{Fu-ESA-2018} and \newcite{Ionescu-KES-2017} by around $4\%$. To our knowledge, our accuracy of $94.1\%$ on RT-2k \cite{Pang-ACL-2004} surpasses all previous results reported in literature. On the MR data set, we surpass most related works by more than $10\%$. To our knowledge, the best accuracy on MR reported in previous literature is $83.6\%$, and it is obtained by \newcite{Cheng-IJCAI-2018}. We surpass the accuracy of \newcite{Cheng-IJCAI-2018} by almost $10\%$, reaching an accuracy of $93.3\%$ using VLAWE. On the TREC data set, we reach the third best performance, after methods such as \cite{Cheng-IJCAI-2018,Zhou-COLING-2016,Zhou-IJCAI-2018}. Our performance on TREC is about $2\%$ lower than the state-of-the-art accuracy of $96.1\%$. On the Subj data set, we obtain an accuracy of $95.0\%$. There are two state-of-the-art methods \cite{Cheng-IJCAI-2018,Zhao-IJCAI-2015} reporting better performance on Subj. Compared to the best one of them \cite{Cheng-IJCAI-2018}, our accuracy is $1\%$ lower. Overall, we consider that our results are noteworthy.

\vspace*{-0.1cm}
\subsection{Discussion}
\vspace*{-0.1cm}

\begin{table}[!t]
\setlength\tabcolsep{5.5pt}
\begin{center}
\begin{tabular}{lccccc}
\hline
Method 										&	MR \\
\hline
\hline
VLAWE ($k=2$) 							& $93.0$ \\
VALWE (PCA) 								& $93.2$ \\										
\hline
VLAWE (full, $k=10$)					& $93.3$ \\
\hline
\end{tabular}
\end{center}
\vspace*{-0.2cm}
\caption{Performance results (in $\%$) of the full VLAWE representation (with $k=10$) versus two compact versions of VLAWE, obtained either by setting $k=2$ or by applying PCA.}
\label{tab_compact}
\vspace*{-0.3cm}
\end{table}

The k-means clustering algorithm and, on some data sets, the cross-validation procedure can induce accuracy variations due to the random choices involved. We have conducted experiments to determine how large are the accuracy variations. We observed that the accuracy can decrease by up to $1\%$, which does not bring any significant differences to the results reported in Table \ref{tab_VLAWE}.

\begin{figure}[!t]
\begin{center}
\includegraphics[width=1.0\linewidth]{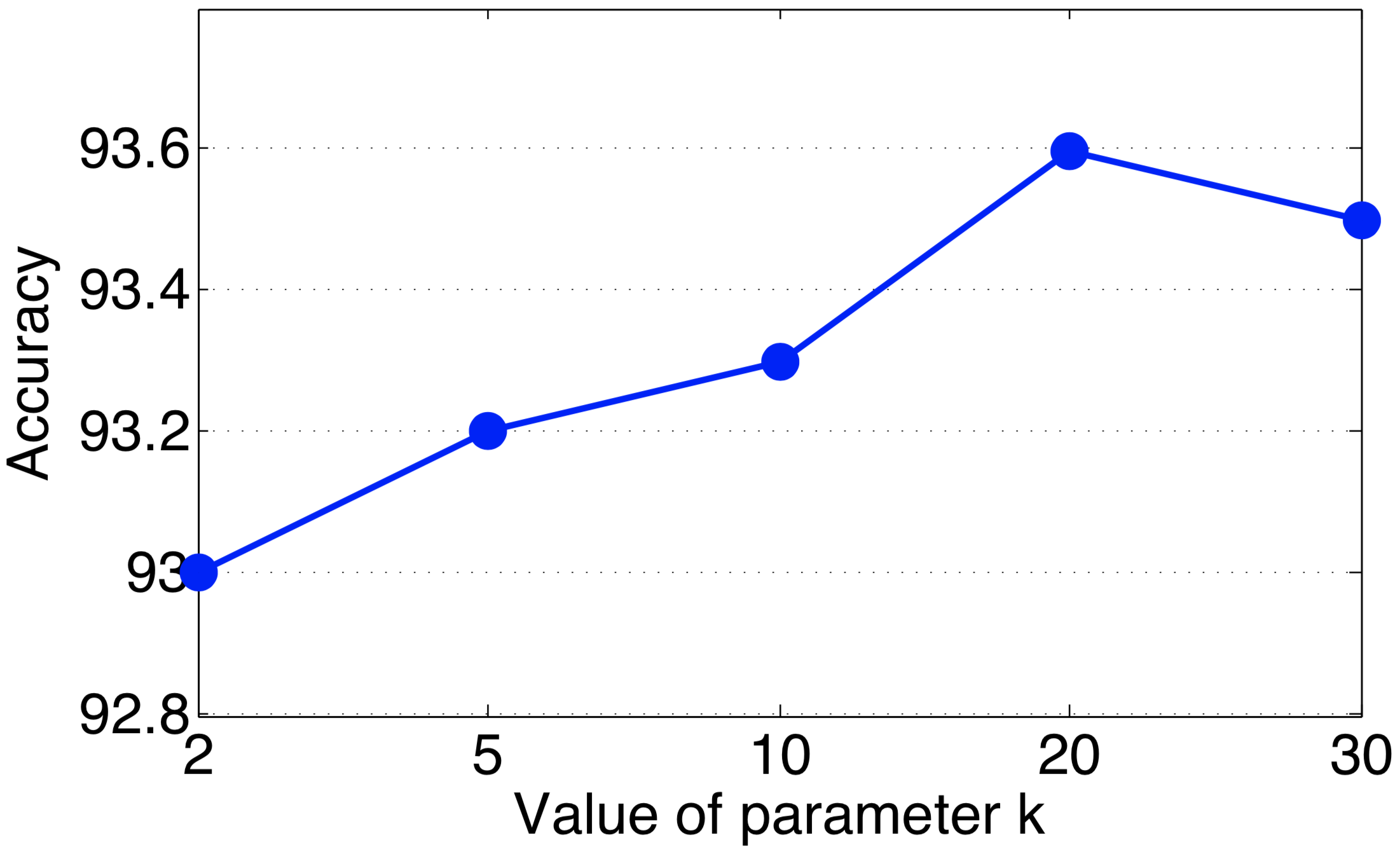}
\end{center}
\vspace*{-0.4cm}
\caption{Accuracy on MR for different numbers of k-means clusters.}
\label{fig1}
\vspace*{-0.5cm}
\end{figure}

Even for a small number of clusters, e.g. $k=10$, the VLAWE document representation can grow up to thousands of features, as the number of features is $k \cdot d$, where $d=300$ is  the dimensionality of commonly used word embeddings. However, there are several document-level representations that usually have a dimensionality much smaller than $k \cdot d$. Therefore, it is desirable to obtain a more compact VLAWE representation. We hereby propose two approaches that lead to more compact representations. The first one is simply based on reducing the number of clusters. By setting $k=2$ for instance, we obtain a $600$-dimensional representation. The second one is based on applying Principal Component Analysis (PCA), to reduce the dimension of the feature vectors. Using PCA, we propose to reduce the size of the VLAWE representation to $300$ components. In Table \ref{tab_compact}, the resulting compact representations are compared against the full VLAWE representation on the MR data set. Although the compact VLAWE representations provide slightly lower results compared to the VLAWE representation based on $3000$ components, we note that the differences are insignificant. Furthermore, both compact VLAWE representations are far above the state-of-the-art method \cite{Cheng-IJCAI-2018}.

In Figure \ref{fig1}, we illustrate the performance variation on MR, when using different values for $k$. We notice that the accuracy tends to increase slightly, as we increase the number of clusters from $2$ to $30$. Overall, the VLAWE representation seems to be robust to the choice of $k$, always surpassing the state-of-the-art approach \cite{Cheng-IJCAI-2018}.

\vspace*{-0.1cm}
\section{Conclusion}
\label{sec:C}
\vspace*{-0.1cm}

We proposed a novel representation for text documents which is based on aggregating word embeddings using k-means and on computing the residuals between each word embedding allocated to a given cluster and the corresponding cluster centroid. Our experiments on five benchmark data sets prove that our approach yields competitive results with respect to the state-of-the-art methods.

\vspace*{-0.1cm}
\section*{Acknowledgments}
\vspace*{-0.1cm}
This research is supported by University of Bucharest, Faculty of Mathematics and Computer Science, through the 2019 Mobility Fund.

\bibliography{references}
\bibliographystyle{acl_natbib}

\end{document}